\documentclass[10pt,twocolumn,letterpaper]{article}

\usepackage{cvpr}              

\usepackage{times}
\usepackage{epsfig}
\usepackage{graphicx}
\usepackage{amsmath}
\usepackage{amssymb}
\usepackage{array}
\usepackage{comment}
\usepackage{booktabs}
\usepackage{multirow}
\usepackage[accsupp]{axessibility}
\usepackage[table]{xcolor}
\newcommand{\maintable} {
\begin{table*}[hbt!]
\begin{center}
\begin{footnotesize}
\begin{tabular}{m{7em} | m{13em} | m{4em}  m{4em} m{4em} m{5em} m{3em}  m{3em}  m{3em} } 
\hline 
\textbf{Datasets} & \textbf{Methods}  & \textbf{Abs Rel}$\downarrow$ &\textbf{ Sq Rel} $\downarrow$ & \textbf{RMSE}$\downarrow$ & \textbf{RMSE(log)}$\downarrow$ & $\pmb{\delta_1}$  $\uparrow$ & $\pmb{\delta_2}$  $\uparrow$ & $\pmb{\delta_3}$  $\uparrow$ \\
\hline \hline
\multirow{6}{*}{Stanford2D3D \cite{armeni2017joint}}  & FCRN \cite{laina2016deeper}  & 0.1837 & - & 0.5774 & -  & 0.7230 & 0.9207 & 0.9731\\
 & RectNet \cite{zioulis2018omnidepth}   & 0.1996& - & 0.6152 & - & 0.6877 & 0.8891 & 0.9578 \\
 & BiFuse with fusion \cite{wang2020bifuse} & 0.1209 & - & 0.4142 & & 0.8660 & 0.9580 & 0.9860\\
 & UniFuse with fusion \cite{jiang2021unifuse} & 0.1114 & - & 0.3691 &  - & 0.8711 & 0.9664 &  0.9882 \\
  & HoHoNet \cite{sun2021hohonet} & 0.1014 & - & 0.3834 & - & \textbf{0.9054} & 0.9693 & 0.9886 \\
\cline{2-9}

 & \textbf{OmniFusion, Ours (1-iter)} & 0.0961 & 0.0543 &	0.3715 &	0.1699 &	0.8940 &	0.9714 &	0.9900 \\
& \textbf{OmniFusion, Ours (2-iter)} &  \textbf{0.0950} &	\textbf{0.0491 }&	\textbf{0.3474} &\textbf{0.1599}	& 0.8988 &	\textbf{0.9769} &	\textbf{0.9924} \\

\hline
\hline
\multirow{8}{*}{Matterport3D \cite{Matterport3D}} & FCRN  \cite{laina2016deeper} & 0.2409  & - & 0.6704 & - & 0.7703 & 0.9714 & 0.9617\\
 & RectNet \cite{zioulis2018omnidepth} & 0.2901 & - & 0.7643 & - & 0.6830 & 0.8794 & 0.9429\\
 & BiFuse with fusion \cite{wang2020bifuse} & 0.2048 & - & 0.6259 & - & 0.8452 & 0.9319 & 0.9632 \\
 & UniFuse with fusion \cite{wang2020bifuse} & 0.1063 & -  & 0.4941 & - & 0.8897 & 0.9623 & 0.9831\\
 & HoHoNet \cite{sun2021hohonet} & 0.1488 & - & 0.5138 & - & 0.8786 & 0.9519 & 0.9771\\ 
\cline{2-9}
 & \textbf{OmniFusion, Ours (1-iter)} & 0.0980 & 0.0611 &	0.4536 & 0.1587	& 0.9040 &	0.9757 &	0.9919\\
  & \textbf{OmniFusion, Ours (2-iter)} & \textbf{0.0900} & \textbf{0.0552} &	\textbf{0.4261} &	\textbf{0.1483} &	\textbf{0.9189} &	\textbf{0.9797} &	\textbf{0.9931} \\
\hline
\hline
\multirow{9}{*}{360D \cite{zioulis2018omnidepth}} & FCRN  \cite{laina2016deeper} & 0.0699  & 0.2833 & & & 0.9532 & 0.9905 & 0.9966\\
 & RectNet \cite{zioulis2018omnidepth}  & 0.0702 & 0.0297 & 0.2911 & 0.1017 & 0.9574 & 0.9933 & 0.9979\\
 & Mapped Convolution \cite{eder2019mapped} & 0.0965 & 0.0371 & 0.2966 & 0.1413 & 0.9068 & 0.9854 & 0.9967 \\
 & BiFuse with fusion \cite{wang2020bifuse} & 0.0615 & - &  0.2440  & - & 0.9699 & 0.9927 & 0.9969\\
  & UniFuse with fusion \cite{wang2020bifuse} & 0.0466 & - & 0.1968 & - & 0.9835 & 0.9965 & 0.9987\\
  & ODE-CNN \cite{cheng2020omnidirectional} & 0.0467 & 0.0124 & \textbf{0.1728} & 0.0793 & 0.9814 & 0.9967 & 0.9989  \\ 
\cline{2-9}
 & \textbf{OmniFusion, Ours (1-iter)} & 0.0469 & 0.0127	& 0.1880	& 0.0792 &	0.9827 &	0.9963	& 0.9988\\
  & \textbf{OmniFusion, Ours (2-iter)} &\textbf{0.0430} & \textbf{0.0114 }&	0.1808 &	\textbf{0.0735}	& \textbf{0.9859} &	\textbf{0.9969} &	\textbf{0.9989} \\
\hline

\end{tabular}
\end{footnotesize}
\caption{Quantitative Results for depth estimation on Stanford2D3d \cite{armeni2017joint}, Matterport3D \cite{Matterport3D}, 360D \cite{zioulis2018omnidepth} datasets. Notably, our method \textit{OmniFusion} achieves state-of-the-art performances in all datasets, outperforming the existing works by a significant margin. 
}
\label{tab:compare1}
\end{center}
\end{table*}

}
\newcommand{\firsttable} {
\begin{table*}[hbt!]
\begin{center}
\begin{footnotesize}
\begin{tabular}{b{24em} b{8em} b{3em}| b{4em}  b{4em} b{4em} } 
\hline 
\textbf{Methods} & \textbf{\#Params} & \textbf{FPS}$\uparrow$ & \textbf{Abs Rel}$\downarrow$ &  \textbf{Sq Rel}$\downarrow$  & \textbf{RMSE}$\downarrow$\\
\hline
Baseline & 23.5M & 9.4 & 0.1136 & 0.0638 & 0.3894 \\
\hline
Baseline + geometric fusion (1-iter) & 23.5M (+1.3K) & 9.3 & 0.1026	& 0.0588	& 0.3812 \\ \hline
Baseline + geometric fusion + transformer (1-iter) & 42.3M (+18.8M)  & 9.2 & 0.0961 &	0.0543	& 0.3715 \\
\hline
\textbf{Baseline + geometric fusion + transformer (2-iter)}& 42.3M (+18.8M)  & 4.6 & \textbf{0.0950} &	\textbf{0.0491}	 & \textbf{0.3474}
\\
\hline

\end{tabular}
\end{footnotesize}
\caption{The ablation study for individual components. Starting from a baseline method with no geometric fusion or transformer, we add each component one at a time. We use ResNet34 for all the experiments. 
}
\vspace{-0.2cm}
\label{tab:ab1}
\end{center}
\end{table*}
}
\newcommand{\secondtable} {
\begin{table}[hbt!]
\begin{center}
\begin{footnotesize}
\begin{tabular}{m{2em} m{4.5em} m{4em} |  m{4em} m{3.5em} m{3.5em}} 
\hline 
\textbf{\#patch} & \textbf{Patch size} & \textbf{Patch FoV}   & \textbf{Abs Rel}$\downarrow$ &  \textbf{Sq Rel}$\downarrow$  & \textbf{RMSE}$\downarrow$ \\
\hline
10 &	256x256 &	120 &	0.1067 &	0.0571 &	0.3788 \\
18 & 128x128 & 80 & 0.1178 &	0.0666	& 0.4018\\
\textbf{18} &	\textbf{256x256}	&   \textbf{80} &	\textbf{0.1037}& \textbf{0.0589}	 & \textbf{0.3686} \\
26 &	256x256	& 60 &	0.1104 &	0.0679 &	0.3955 \\
46 &	128x128	& 50 &	0.1181 &	0.0680 &	0.4101 \\
\hline

\end{tabular}
\end{footnotesize}
\caption{The ablation study for patch size and number of patches. }
\label{tab:ab2}
\end{center}
\end{table}
}
\newcommand{\thirdtable} {
\begin{table}[hbt!]
\begin{center}
\begin{footnotesize}
\begin{tabular}{b{4em} b{2em}  b{2em}| b{4em}  b{4em} b{4em}} 
\hline 
\textbf{Encoder} & \textbf{\#iters} & \textbf{FPS}$\uparrow$ & \textbf{Abs Rel}$\downarrow$ &  \textbf{Sq Rel}$\downarrow$  & \textbf{RMSE}$\downarrow$ \\
\hline
ResNet18 & 1 &\textbf{9.8} & 0.1037 &	0.0589 & 0.3686 \\
ResNet18 & 2 & 4.6 & 0.0979 &	0.0539 & 0.3702 \\
ResNet18 & 3 & 3.1 &\textbf{0.0981} & 0.0521 & \textbf{0.3699}\\
ResNet18 & 4 & 1.5 &  0.0983 & \textbf{0.0519} & 0.3700    \\
\hline
ResNet34 & 1 & \textbf{9.2} & 0.0961 &	0.0543 & 0.3715\\ 
ResNet34 & 2 & 4.6 & 0.0950 &	0.0491 & \textbf{0.3474}\\ 
ResNet34 & 3 & 2.9 & \textbf{0.0894} & \textbf{0.0482} & 0.3498 \\ 
ResNet34 & 4 & 1.4  & 0.0899 & 0.0485 & 0.3491  \\ 
\hline

\end{tabular}
\end{footnotesize}
\caption{The ablation study for different encoder models and different number of iterations.}
\label{tab:ab3}
\vspace{-0.5cm}
\end{center}
\end{table}
}
\usepackage[pagebackref,breaklinks,colorlinks]{hyperref}

\newcommand{\todo}{\textcolor{red}}

\usepackage[capitalize]{cleveref}
\crefname{section}{Sec.}{Secs.}
\Crefname{section}{Section}{Sections}
\Crefname{table}{Table}{Tables}
\crefname{table}{Tab.}{Tabs.}

\def\cvprPaperID{3845} 
\def\confName{CVPR}
\def\confYear{2022}

\begin{document}

\title{OmniFusion: 360 Monocular Depth Estimation via Geometry-Aware Fusion} 

\author{
	Yuyan Li$^{1}$\thanks{Equal contribution} \hspace{10pt} Yuliang Guo$^{2*}$ \hspace{10pt} Zhixin Yan$^{2}$ \hspace{10pt} Xinyu Huang$^{2}$
	\hspace{10pt} Ye Duan$^{1}$
	\hspace{10pt}Liu Ren$^{2}$\\
	\vspace{-0.15cm}
	\small{$^{1}$University of Missouri} 
	\hspace{5pt}
	\small{$^{2}$Bosch Research North America}\\
	\vspace{-0.15cm} \\
	{\tt\small \{yl235, duanye\}@umsystem.edu \hspace{3pt}  
	\{yuliang.guo2, zhixin.yan2, xinyu.huang, liu.ren\}@us.bosch.com}\\
	\vspace{-0.7cm}
}
\maketitle

\begin{abstract}
A well-known challenge in applying deep-learning methods to omnidirectional images is spherical distortion. 
In dense regression tasks such as depth estimation, where structural details are required, using a vanilla CNN layer on the distorted 360 image results in undesired information loss.
In this paper, we propose a 360 monocular depth estimation pipeline, \textit{OmniFusion}, to tackle the spherical distortion issue. 
Our pipeline transforms a 360 image into less-distorted perspective patches (i.e. tangent images) to obtain patch-wise predictions via CNN, and then merge the patch-wise results for final output.
To handle the discrepancy between patch-wise predictions which is a major issue affecting the merging quality, we propose a new framework with the following key components.
First, we propose a geometry-aware feature fusion mechanism that combines 3D geometric features with 2D image features to compensate for the patch-wise discrepancy. 
Second, we employ the self-attention-based transformer architecture to conduct a global aggregation of patch-wise information, which further improves the consistency. 
Last, we introduce an iterative depth refinement mechanism, to further refine the estimated depth based on the more accurate geometric features. 
Experiments show that our method greatly mitigates the distortion issue, and achieves state-of-the-art performances on several 360 monocular depth estimation benchmark datasets. Our code is available at \url{https://github.com/yuyanli0831/OmniFusion}.
\end{abstract}

\section{Introduction}
\label{sec:intro}

\begin{figure}[t]
    \centering
    \includegraphics[width=0.475\textwidth]{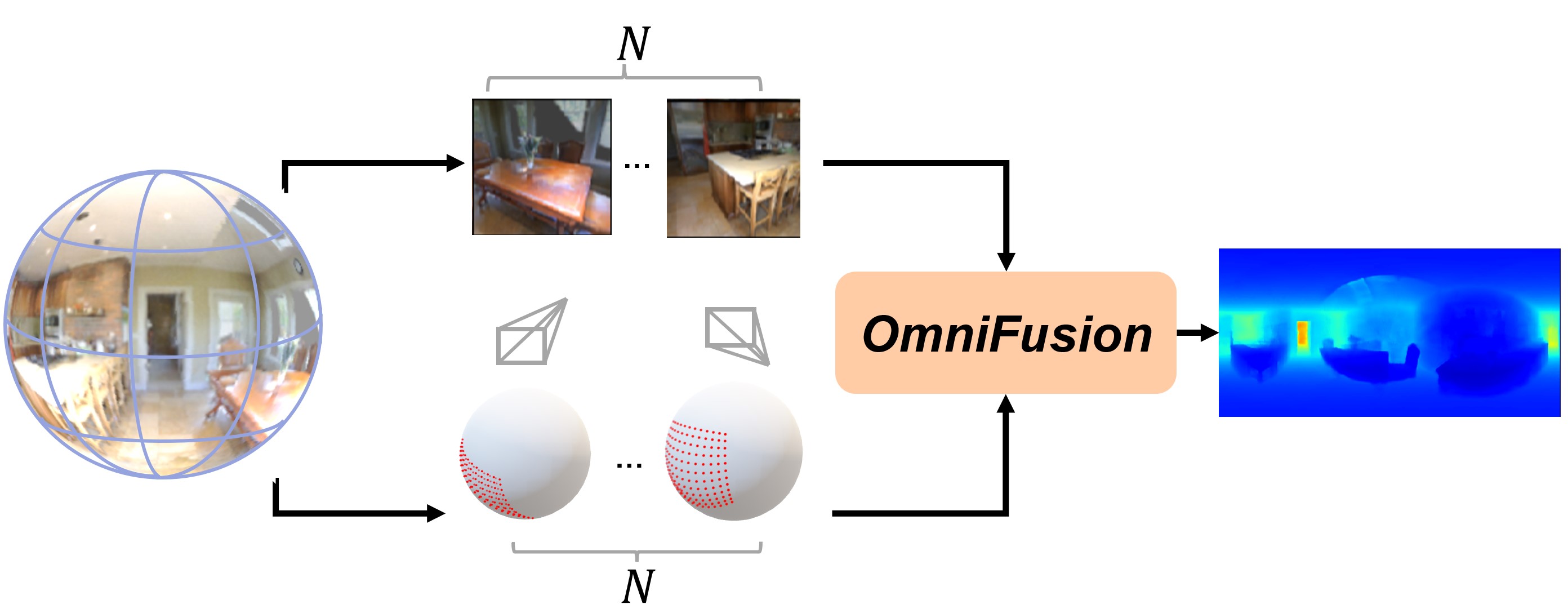}
    \caption{Our method, \textit{Omnifusion}, produces high-quality dense depth (shown as the image on the right) from a monocular ERP input (shown as an image wrapped on a unit sphere on the left). Our method uses a set of $N$ perspective patches (i.e. tangent images) to represent the ERP image (top branch), and fuse the image features with 3D geometric features (bottom branch) to improve the estimation of the merged depth map. The corresponding camera poses of the tangent images are shown in the middle row. }
    \label{fig:overview}
\end{figure}

A 360 image provides a comprehensive view of the scene with its wide field of view (FoV), which is beneficial in understanding the scene holistically. 
However, commonly used 360 image representation format such as the equirectangular projection (ERP) image can introduce geometric distortions. The distortion factor varies in the vertical direction and may degrade the performance of regular convolutional layers designed for non-distorted perspective images.
Many studies have been proposed to address the distortion issue. \cite{su2019kernel,chen2021distortion,zhao2018distortion} proposed distortion-aware convolutions or spherical customized kernels. However, it remains unclear how effective such spherical convolutions are, especially in deeper layers \cite{wang2020bifuse, su2019kernel}. Some spherical CNNs \cite{cohen2018spherical, yang2020panoramic} defined convolution in the spectral domain, with potentially heavier computation overhead. Attempts have also been made to tackle the ERP distortion via other less-distorted formats.
BiFuse \cite{wang2020bifuse} and UniFuse \cite{jiang2021unifuse} took complementary properties from ERP and cubemap. Several works \cite{chou2018self, su2016pano2vid} applied regular CNNs repeatedly to multiple perspective projections of the 360 image.
Recently, Eder et al. \cite{eder2020tangent} proposed to use a set of subdivided icosahedron tangent images, and demonstrated that using tangent image representation can facilitate the network transfer between perspective and 360 images. 

It is advantageous to use tangent images \cite{eder2020tangent} as it has less distortion, and can make good use of the large pool of pre-trained CNNs developed for perspective imaging. Additionally, the tangent image representation inherits a superior scalability to handle high resolution inputs compared to those holistically method. 
However, the vanilla pipeline \cite{eder2020tangent} has some limitations.
First, severe discrepancies occur between perspective views since the same object may appear differently from multiple views (an example is shown in Figure \ref{fig:projection}). This issue is especially problematic for the depth regression task, since the inconsistent depth scale estimated from individual tangent images creates undesired artifacts during merging.
Second, the advantage of estimating depth from holistic 360 image is unfortunately lost, because of the decomposition of the global scene into local tangent images. The predictions from the tangent images are independent of each other and there is no information exchange between tangent images. 

In this paper, we present \textit{OmniFusion}, a 360 monocular depth estimation framework with geometry-aware fusion (see Figure \ref{fig:overview}). We proposed the following three key components to solve the aforementioned discrepancy issue and merge the depth results of tangent images seamlessly.
First, we use a geometric embedding module to provide additional features to compensate for the discrepancy between 2D features from patch to patch. For each patch, we calculate the 3D points located on the spherical surface that correspond to the patch pixels, encode them and the patch center coordinate through shared Multi-layer Perceptron (MLP), and add the geometric features to the corresponding 2D features. 
Second, to regain the holistic power in understanding the entire scene,  we incorporate a self-attention-based transformer in our pipeline. With the transformer, patch-wise information is globally aggregated to enhance the estimation of the global scale of depth, and to improve the consistency between patch-wise results. 
Third, we introduce an iterative refining mechanism, where more accurate 3D information from the predicted depth maps is fed back to the geometric embedding module to further improve the depth quality in an iterative manner. 

We test \textit{OmniFusion} on three benchmark datasets: Stanford2D3D \cite{armeni2017joint}, Matterport3D \cite{Matterport3D}, and 360D \cite{zioulis2018omnidepth}. Experimental results show that our method outperforms state-of-the-art methods by a significant margin on all of these datasets. 

Our contributions can be summarized as follows:
\begin{itemize}
    \item We present a 360 monocular depth prediction pipeline that addresses the distortion issue via geometry-aware fusion and achieves the state-of-the-art performance.
    \item We introduce a geometric embedding network to provide 3D geometric features to mitigate the discrepancy in patch-wise image features.
    \item 
    We incorporate a self-attention-based transformer to globally aggregate patch-wise information which enhances the estimation of the physical scale of depth.
    \item We propose an iterative mechanism to further improve the depth estimation with structural details.

\end{itemize}

\section{Related Work}
\label{sec:related}

\subsection{Monocular depth estimation}

Monocular depth estimation, which takes a single RGB image as input to predict pixel-wise depth value, 
has been extensively investigated due to its broad applications. Early works mainly focused on network architecture and supervision~\cite{eigen2014depth,laina2016deeper,he2016deep}.
Recently, researchers has been investigating the use of unsupervised learning on stereo pairs~\cite{GargKC0:eccv16,XieGF:eccv16,godard2017unsupervised} or monocular video streams~\cite{zhou2017unsupervised,godard2019digging} to expand training data to unlabelled image sequences for broader applications.
However, such approaches are still sensitive to many factors (e.g. camera intrinsic changes), and very challenging to be generalizable to new scenes. 
To improve the robustness and scalability, some methods utilize  additional  sensor input such as LiDAR and RGBD camera~\cite{LinDG:iros20,cheng2020omnidirectional}. 
However, the extra computation or power consumption are not welcomed in many practical scenarios. 

\begin{figure*}[t]
    \centering
    \includegraphics[width=17cm]{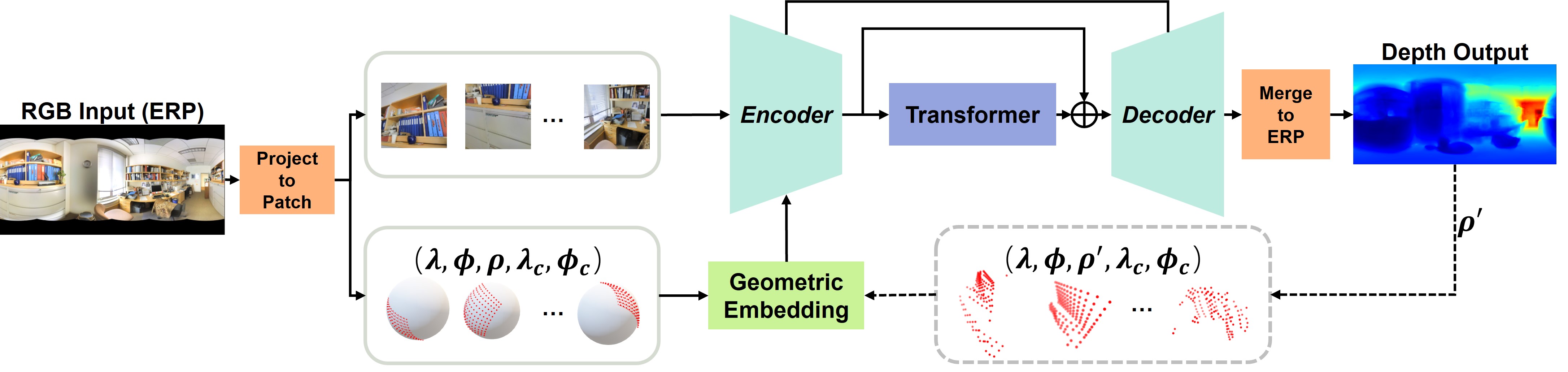}
    \caption{Overview of our proposed \textit{OmniFusion}. Our method takes a monocular ERP RGB as input, projects it onto multiple patches at multiple viewpoints, and processes each distortion-free patch with an encoder-decoder network to produce patch-wise depth maps (top-stream). The patch-wise outputs are merged into a final ERP depth map. Meanwhile, the corresponding points located on the spherical surface are sampled and passed through a geometric embedding network to produce geometric features (bottom-stream). The geometric features are fused into the image encoder to compensate for the patch-wise discrepancy and to improve  the quality of the merged result. For each sampled point, we use its spherical coordinates $(\lambda, \phi, \rho)$, together with the tangent plane center coordinates $(\lambda_c, \phi_c)$ as input attributes to the geometric embedding network which provides the necessary information to align 2D features. A transformer architecture is integrated to conduct global aggregation of the deep patch-wise feature which further improves the consistency of patch-wise outputs. 
    Moreover, we incorporate an iterative refining mechanism (visualized in dashes), to further improve the depth recovery. In particular, $\rho$ value is updated according to the depth estimated from the previous iteration. }
    \label{fig:pipeline}
\end{figure*}

\subsection{360 depth estimation}
Monocular depth estimation from 360 images has been investigated from a variety of perspectives.  Zioulis et al. \cite{zioulis2019spherical} explored the spherical stereo geometry and estimated depth from monocular ERP input via view synthesis. PanoDepth \cite{li2021panodepth} leveraged 360 stereo constraints to improve monocular depth performance.
Eder et al.\cite{eder2019pano} and Zeng et al.\cite{zeng2020joint} explored joint learning from different modalities (e.g. layout, normal, semantics, etc.). HoHoNet \cite{sun2021hohonet} proposed to utilize latent horizontal feature representation to encode ERP image features. 
To handle the irregular distortion of ERP images, several distortion-aware convolutions \cite{su2017learning,fernandez2019CFL,chen2021distortion,zhao2018distortion,su2019kernel} have been proposed. For example, Fernandez et al. \cite{fernandez2019CFL} introduced EquiConv 
which applied deformable convolution to accommodate spherical geometry. 
Tateno et al.\cite{Tateno_2018_ECCV} proposed to apply regular CNN to perspective images during training, and distortion-aware convolution during testing. Instead of directly tackling the distortion of ERP, several approaches proposed to use other representations with less distortion, such as cubemap \cite{wang2018self, cheng2018cube}, fusion between ERP and cubemap \cite{wang2020bifuse, jiang2021unifuse}, and multiple perspective projections of 360 images\cite{chou2018self, su2016pano2vid}.
A recent work by Eder et al. \cite{eder2020tangent} proposed to use tangent images, a set of oriented, low-distortion images rendered tangent to faces of the icosahedron, to represent a 360 image. 
It is advantageous to use tangent images since it has less distortion and can effectively leverage pre-trained CNN models developed for perspective imaging.
However, discrepancies between tangent images are not addressed in \cite{eder2020tangent}, which leads to a downgrade of the final merged result. In this work, we follows the paradigm proposed in \cite{eder2020tangent} of using tangent images, but simplified and adapted it for depth estimation.  In addition, we successfully address the discrepancy issue by incorporating geometry-aware fusion and the transformer.


\subsection{Transformer}
Originally proposed in natural language processing \cite{vaswani2017attention}, the transformer architecture has since been widely used in computer vision tasks such as image classification \cite{dosovitskiy2020image}, depth estimation \cite{Ranftl_2021_ICCV}, object detection \cite{carion2020end}, and semantic segmentation \cite{strudel2021segmenter,zheng2021rethinking}.
The visual transformer has a natural fit with monocular depth estimation as long-range context can be explicitly exploited by the self-attention module. 
When applying transformer to 360 images, the distortion however, can decrease the power of the transformer in exploiting the pairwise correlation between patches. In this work, we feed the transformer with distortion-free, and geometry-aware input, so that the transformer can focus on the global aggregation of patch-wise information.

\section{Method}
\label{sec:method}
Figure \ref{fig:pipeline} shows an overview of the full pipeline of the proposed \textit{OmniFusion} framework. First, an ERP input image is transformed into a set of tangent images via gnomonic projection (Figure \ref{fig:projection}). The projected distortion-free tangent images are then passed through an encoder-decoder network to produce patch-wise depth estimations, which are later fused into an ERP depth output. To ease the patch-wise discrepancy, we introduce a novel geometric embedding module that encodes the spherical coordinate associated with each tangent image pixel, providing additional geometric features to facilitate the integration of patch image features. To further improve the consistency between patch-wise predictions and to better estimate the global depth scale, the features from the deepest level of the encoder are globally aggregated through a self-attention-based transformer. Finally, an iterative refining mechanism is adopted to further improve the depth quality. We update the spherical coordinates iteratively based on the more accurate estimation obtained from the previous iteration. We train our network in an end-to-end fashion, with the only supervision being the final merged depth compared to the ground truth.

\begin{figure}[t]
    \centering
    \includegraphics[width=7.5cm]{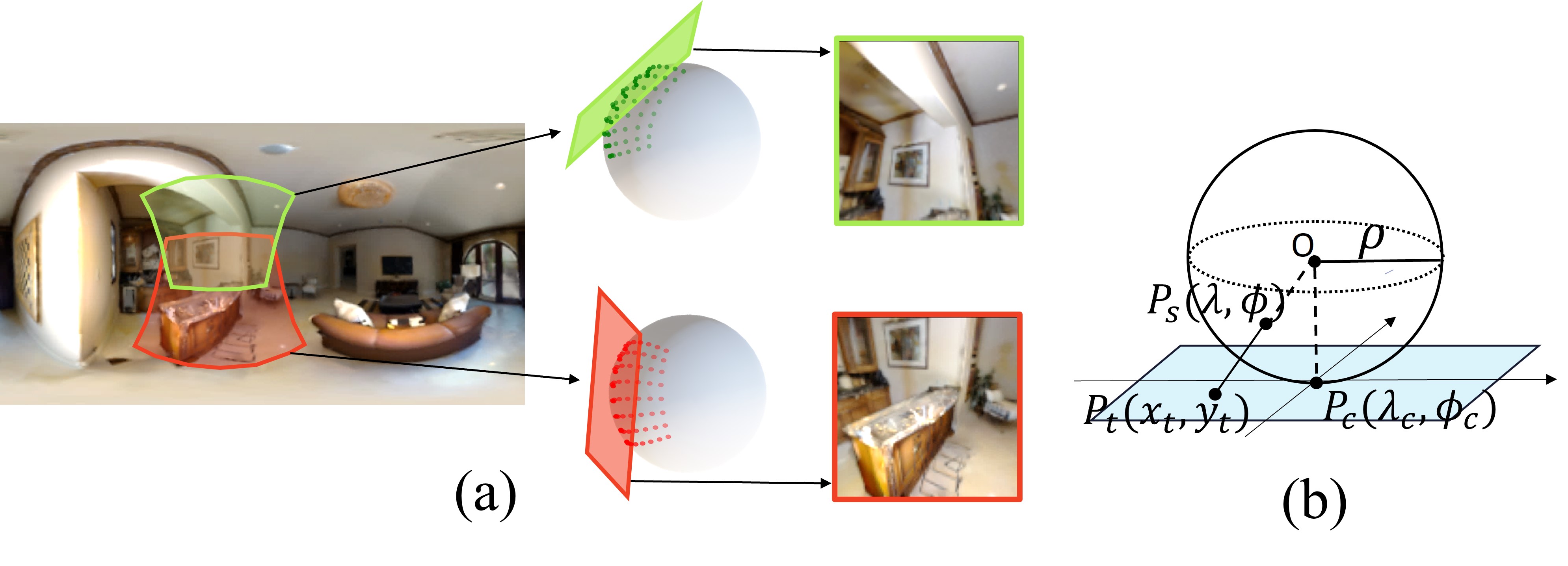}
    \caption{(a) An example of tangent image projection. Two tangent images are projected  from two different viewpoints. The corresponding areas are highlighted with the same color in both ERP and tangent patches. As illustrated, there usually exist overlapping areas between two neighboring patches, and the same object may appear differently in different patches. (b) The illustration of the gnomonic projection. A point $P_s(\lambda, \phi)$ located on the spherical sphere is projected onto a point $P_t(x_t, y_t)$ on the flat plane which is tangent to a point $P_c(\lambda_c, \phi_c)$. 
     }
    \label{fig:projection}
\end{figure}

\begin{figure*}[t]
    \centering
    \includegraphics[width=14cm]{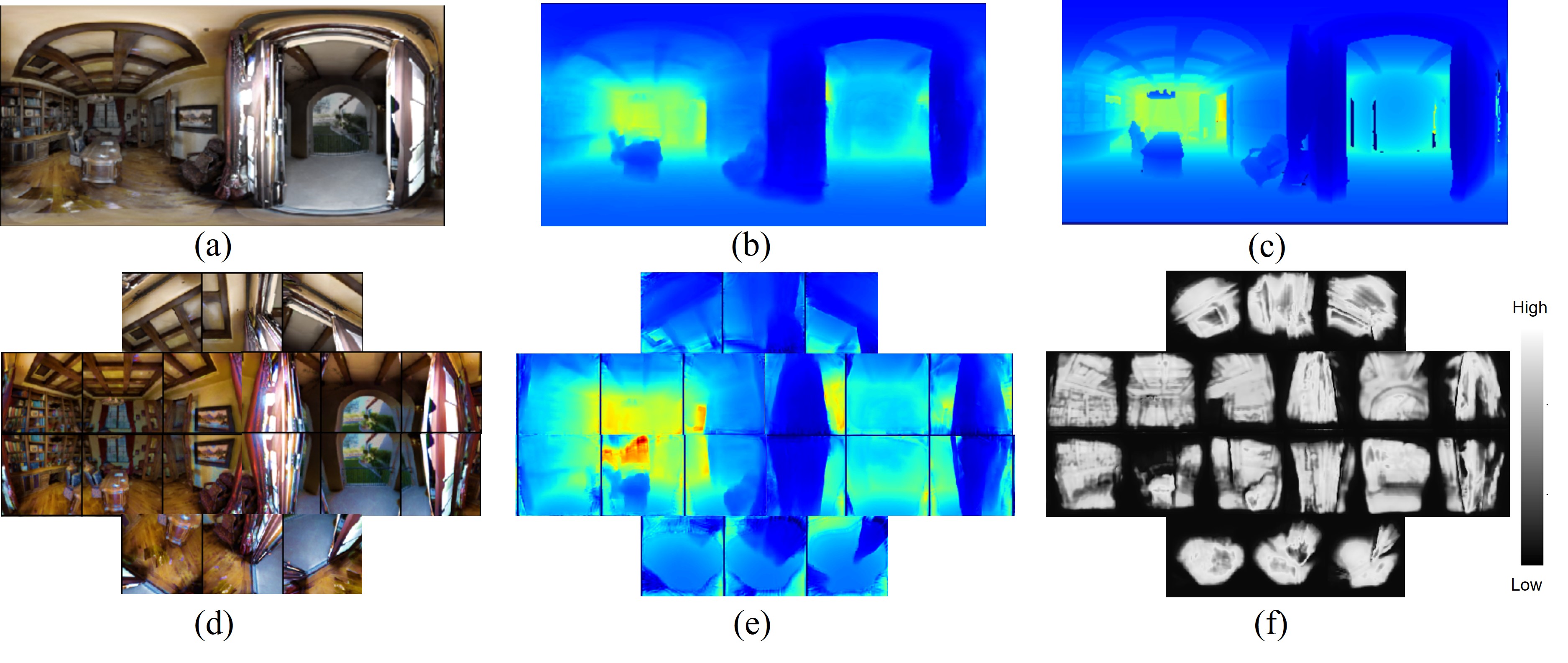}
    \caption{
    First row: (a) An example of an ERP RGB input image, (b) the final merged predicted ERP depth map, (c) the ground truth ERP depth. Second row: (d) RGB tangent image patches generated from (a), (e) the patch-wise estimated depth maps, (f) the patch-wise estimated confidence maps that are used as weights and facilitates the ERP depth merging. }
    \label{fig:exp1}
\end{figure*}

\subsection{Depth estimation from tangent images}
\label{sec:baseline}

Our method relies on the less-distorted tangent images to address the irregular distortion in 360 images. A tangent image is generated via \textit{gnomonic projection} of a sphere surface onto a flat, rectangular plane surface. The gnomonic projection \cite{coxeter1961introduction} (Figure \ref{fig:projection}) is a mapping obtained by projecting points $P_s (\lambda, \phi)$ on the surface of sphere from a sphere's center $O$ to point $P_t (x_t, y_t)$ in a plane that is tangent to a point $P_c (\lambda_c, \phi_c)$. We use $(\lambda, \phi)$ to indicate the longitude and latitude, respectively, and $(x_t, y_t)$ to indicate a 2D point position on the tangent image. The detailed formulas are included in the Appendix.

In our experiments, we use a set of $N=18$ tangent images for a balance of speed and performance (A related ablation study can be found in Section \ref{sec:ablation}). Tangent images are sampled at four different latitudes: $-67.5^\circ, -22.5^\circ, 22.5^\circ, 67.5^\circ$, and  we sample $3, 6, 6, 3$ patches on each of these latitudes, respectively (Figure \ref{fig:exp1}). All tangent images share the same resolution and FoV. We chose this non-uniform sampling based on the fact that tangent images of the same resolution can cover different ranges of longitude when the centered at different latitudes. 
To ensure the sampled patches near the poles do not overlap to an extreme extent, we take fewer samples to cover the near-pole area in the ERP space. 
Since the generated tangent images are distortion-free, we can easily apply regular encoder-decoder CNN architectures to predict a depth map from each tangent image. For better convergence and accuracy, we leverage high-performance pre-trained networks (e.g., ResNet \cite{he2016deep}) when initializing our encoder. We pass all $N$ tangent images simultaneously through the encoder, and obtain $N$ feature maps that will be used as tokens later in the transformer.
For the decoder, we use a stack of upsampling layers followed by $3\times 3$ convolutions, with skip-connections from the encoder.

The baseline presented so far can be considered as a customized version of \cite{eder2020tangent}. We adopt different rendering of tangent images and the network architecture from \cite{eder2020tangent} to make the baseline method more effective and efficient. Note that for our baseline, no transformer, geometric fusion, or confidence map is used, the output depth is the average of all patches.





\subsection{Geometry-aware feature fusion}
\label{sec:geo_emb}

\begin{figure}[t]
    \centering
    \includegraphics[width=8.2cm]{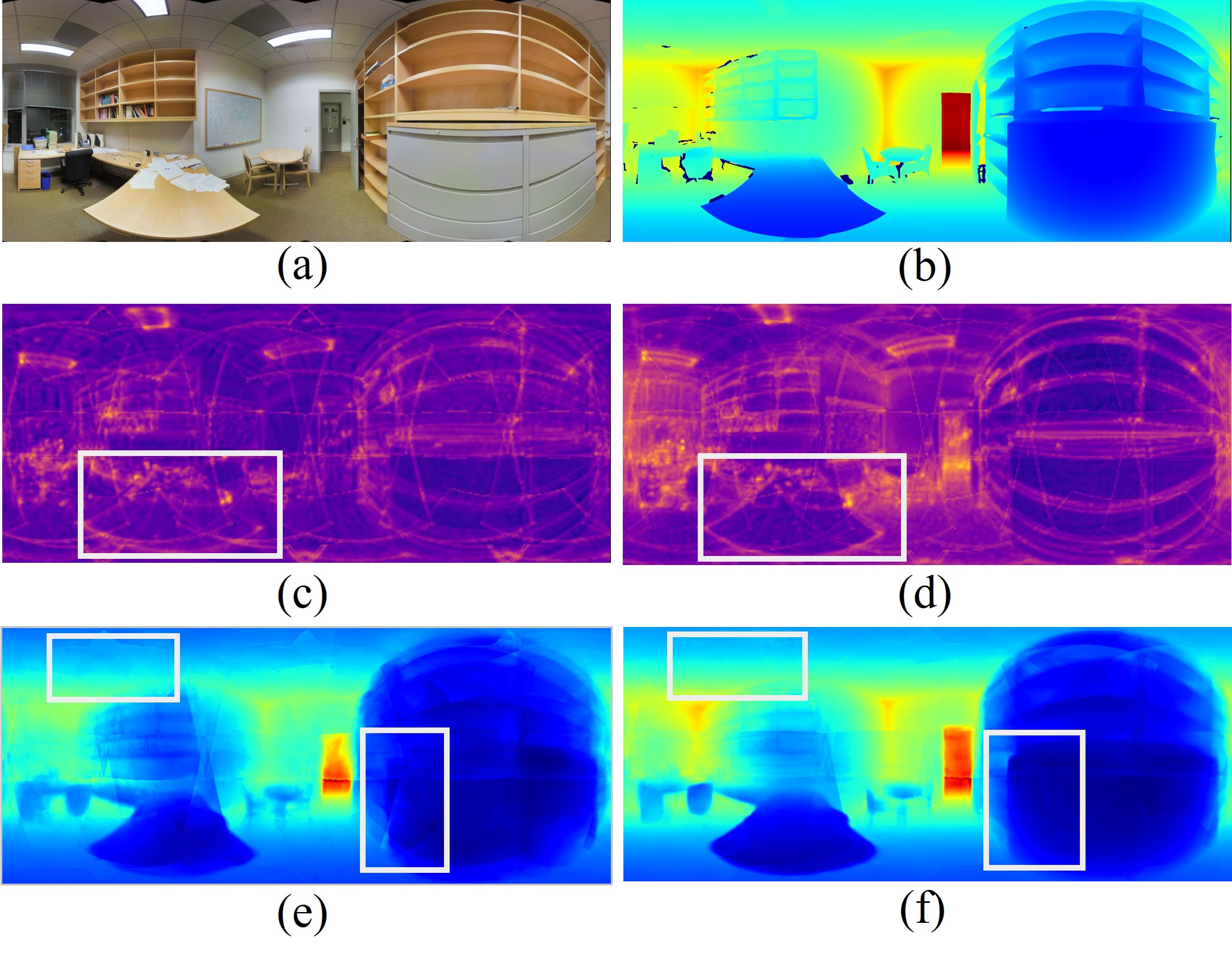}
    \caption{Illustration of the effectiveness of geometry-aware feature fusion. An ERP RGB image is shown in (a), the ground truth depth is shown in (b). Visualizations of the feature map and the final depth map from the baseline are shown in (c) and (e) respectively. For comparison, (d) and (f) show the feature map and the final depth map out of the proposed \textit{OmniFusion}, where the image features are fused with geometric features. Observe that our method yields a more self-consistent feature map and a more structural depth map compared to the baseline, especially in regions highlighted in rectangles.}
    \label{fig:geo:embed:effect}
    \vspace{-0.5cm}
\end{figure}

The simplicity of predicting depth maps from tangent images nonetheless comes with a cost. As the depth estimation is now conducted independently, a globally consistent depth scale is no longer guaranteed. Furthermore, as shown in Figure~\ref{fig:projection} (a) and Figure~\ref{fig:exp1} (d), an object (e.g. the painting on the wall in Figure~\ref{fig:projection} (a)) will be projected onto multiple tangent images from various angles and therefore will be encoded differently in different tangent images. Discrepancies between patch depth estimations, especially in overlapping areas, can result in significant artifacts in the final merged ERP depth map (Figure~\ref{fig:geo:embed:effect} (e)).

To compensate for the differences between patch-wise image features, we introduce a \textit{geometric embedding} network (see Figure~\ref{fig:pipeline}) to provide additional geometric information. For a pixel $P_t(x_t, y_t)$ located on a tangent image, we use its corresponding spherical coordinates located on the unit sphere, $P_s(\lambda, \phi, \rho)$, together with the center of the tangent image $P_c(\lambda_c, \phi_c)$, as the input attributes of the geometric embedding network. $P_s$ makes the embedding aware of the global position, e.g., to tell whether two image pixels from two patches relate to the same spherical coordinates. However, geometric features out of $P_s$ alone can not align different 2D features. To this end, $P_c$ is taken as additional attributes to make the embedding able to differentiate from patch to patch, such that the learned geometric features can make the patch features tend to be consistent.
Through the combination of the tangent image features and the geometric features as well as an end-to-end learned network, the adjusted features lead to a much cleaner merged depth.
As observed in Figure~\ref{fig:geo:embed:effect} (d), the extracted image features with geometric embedding show much better consistency in the feature map merged in the ERP space, compared to features without geometric embedding as shown in Figure~\ref{fig:geo:embed:effect} (c). Consequently, the final depth map out of \textit{OmniFusion} shown in Figure~\ref{fig:geo:embed:effect} (f) appears to be much cleaner compared to the baseline depth map shown in Figure~\ref{fig:geo:embed:effect} (e). 

The geometric embedding network consists of two layers of MLPs, and encodes the 5-channel spherical attributes into 64-channel feature maps. We fuse this geometric embedding with image features at the same pixel location in the encoder via element-wise summation. In order to maintain more structural details, early fusion is adopted. Geometric features are added to the \textit{layer1} of the ResNet encoder where we experimentally achieved the best performance. It is worth noting that the additional computational cost associated with the geometric embedding module is minimal compared to the original encoder-decoder (Table \ref{tab:ab1}). The geometric features for the first iteration are fixed once learned, since they are independent from image inputs. Only the second iteration requires to re-compute the geometric features.

\subsection{Global aggregation with transformer}
When decomposing the ERP into a sequence of tangent images, we no longer have the holistic view of the 3D environment. To make up for this loss,
we leverage the transformer architecture to aggregate information from the patches in a global fashion. The global aggregation is expected to improve the consistency of depth estimations from patches, and to better regress the global scale of depth out of a larger FoV. 

Using the feature maps extracted from the encoder, we first apply a $1\times 1$ convolution layer to reduce channel dimensions for better efficiency. Then we flatten the feature maps into $N$ 1-D feature vectors $X_0 = [x^1, x^2, ..., x^N] \in R^{N\times d}$ which will be used as tokens in the transformer. The learnable positional embedding $E_{pos} \in R^{N\times d}$ are added to the feature tokens to retain positional information in a similar way as proposed in \cite{dosovitskiy2020image}. Through the self-attention architecture, the transformer learns to globally aggregate the information from all the patches to adjust the features from each patch, where the aggregation weights account for the pairwise correlation both from the visual features and the positional features. The architecture of the multi-head attention transformer follows \cite{vaswani2017attention}. 

\subsection{Depth merging with learnable confidence map}
\label{sec:depth:fusion}

The aforementioned geometric embedding and transformer modules significantly reduce discrepancies among different patch-wise depth estimations. 
Yet, the depth merging does not achieve a pixel-level seamless fusion.
To further improve the merging (Figure~\ref{fig:exp1} (b)), we ask the network to simultaneously predict a confidence map for each patch besides depth regression. The merged depth is then computed as a weighted average of all patch depth predictions with confidence scores used as weights. In detail, two separate regression layers are appended to the decoder, one for depth regression, the other for confidence score regression.
Both the depth maps (Figure~\ref{fig:exp1} (e)) and confidence maps (Figure~\ref{fig:exp1} (f)) are mapped to ERP domain following the inverse gnomonic transformation before merging. (More  details are included in the Appendix.)


\subsection{Iterative depth refinement}
The geometric embedding utilizes the spherical coordinates $(\lambda, \phi, \rho)$ corresponding to tangent image pixels for geometry-aware fusion. $\rho$ is initially fixed as no depth information is available. The depth information will be available after one iteration, which can be used to update $\rho$ and provide more accurate geometry information for the geometric embedding module. Based on this observation, we propose an iterative depth refinement scheme (see Figure \ref{fig:pipeline}).

In the first iteration (Section \ref{sec:geo_emb}), the spherical coordinates $(\lambda, \phi, \rho)$ of points located on the unit sphere are used for geometric embedding. For the subsequent iterations, we update $\rho \rightarrow \rho'$, using the new depth value estimated from the previous iteration (the depth of ERP image is defined as the distance from the real-world point to the camera center). The updated attributes with more accurate geometry will be passed into the geometric embedding network in the next iteration. 
An ablation study is presented in Section \ref{sec:exp} to demonstrate the effectiveness of more accurate geometric embedding.

\maintable

 \begin{figure*}[htb!]
     \centering
     \includegraphics[width=15.5cm]{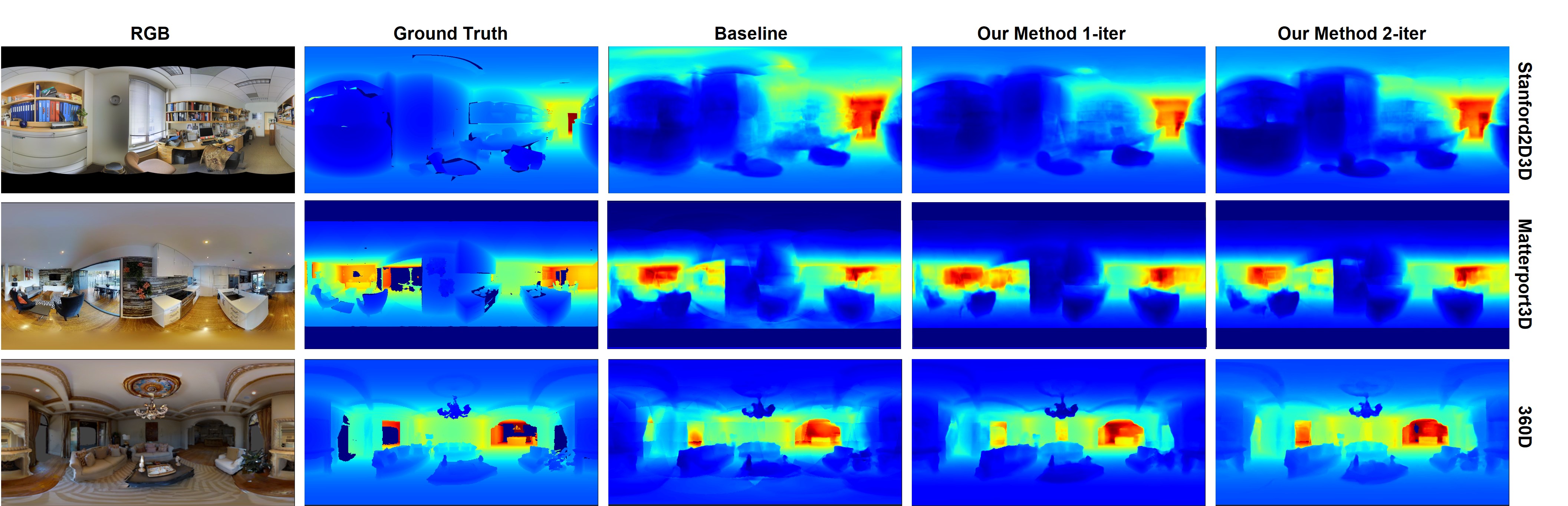}
     \caption{Qualitative results on Stanford2D3D \cite{armeni2017joint}, Matterport3D \cite{Matterport3D} and 360D \cite{zioulis2018omnidepth}. From left to right: ERP RGB input, ground truth depth, depth output from the baseline, depth output from our method 1-iter and 2-iter.
     In comparison to the baseline (described in Section \ref{sec:baseline}, our method (1-iter, 2-iter) leads to more structural depth, which appear sharp along those object boundaries and smooth within surfaces.
     }
     \label{fig:quali}
 \end{figure*}
 
\section{Experiments}
\label{sec:exp}

\subsection{Datasets}
\label{sec:dataset}
\textit{OmniFusion} is tested on three well-known benchmark datasets: Stanford2D3D \cite{armeni2017joint},  Matterport3D \cite{Matterport3D}, 360D \cite{zioulis2018omnidepth}.
\noindent\textbf{Stanford2D3D \cite{armeni2017joint}} dataset consists of 1,413 real world panorama images from six large-scale indoor areas. We follow the official train-test split which uses the fifth area for testing, and others for training. We use  resolution $512 \times 1024$.

\noindent\textbf{Matterport3D \cite{Matterport3D}} contains a total of 10,800 indoor panorama RGBD images. We follow the official split which takes 61 rooms for training and the rest for testing. We use resolution $512\times 1024$ in our experiments.

\noindent\textbf{360D \cite{zioulis2018omnidepth}} is a RGBD panorama benchmark provided by Zioulis et al. \cite{zioulis2018omnidepth}. It is composed of two other synthetic datasets (SunCG and SceneNet), and two real world datasets (Stanford2D3D and Matterport3D). There are 35,977 photo-realistic panorama RGBD images in the 360D that are rendered from the aforementioned four datasets. We follow the default train-test splits and use resolution $256\times 512$. 

\subsection{Implementation details}
We adopt the same quantitative evaluation metrics as used in \cite{laina2016deeper, zioulis2018omnidepth}, including Absolute Relative Error (Abs Rel), Root Mean Squared Error (RMSE), Root Mean Squared Error in logarithmic space (RMSE(log)) and accuracy with a threshold $\delta_t$, where $t \in {1.25,1.25^2,1.25^3}$. Arrows next to the metric indicate the direction of better performance in all tables. 
We implement our network using PyTorch and train it on two Nvidia RTX GPUs. We use the default setting of Adam optimizer \cite{kingma:adam} and a initial learning rate of 0.0001 with cosine annealing \cite{loshchilov2016sgdr} learning rate policy. We train Stanford2D3D \cite{armeni2017joint} for 80 epochs, and 60 epochs for Matterport3D \cite{Matterport3D} and 360D \cite{zioulis2018omnidepth}. The default number of patches we use is 18. The default patch size we use for Stanford2D3D \cite{armeni2017joint} and matterport \cite{Matterport3D} is $256\times 256$, the patch FoV is $80^{\circ}$. For 360D \cite{zioulis2018omnidepth}, we use $128\times 128$ as patch size. We leverage pre-trained ResNet \cite{he2016deep} as image encoder in these experiments.
The network is trained end-to-end, the same model is used for all iterations. For the loss function, following  \cite{Tateno_2018_ECCV,wang2020bifuse}, we adopt BerHu loss \cite{laina2016deeper} for depth supervision. The final loss is the summation of depth losses from all iterations.

\begin{figure*}
    \centering
    \includegraphics[width=16cm]{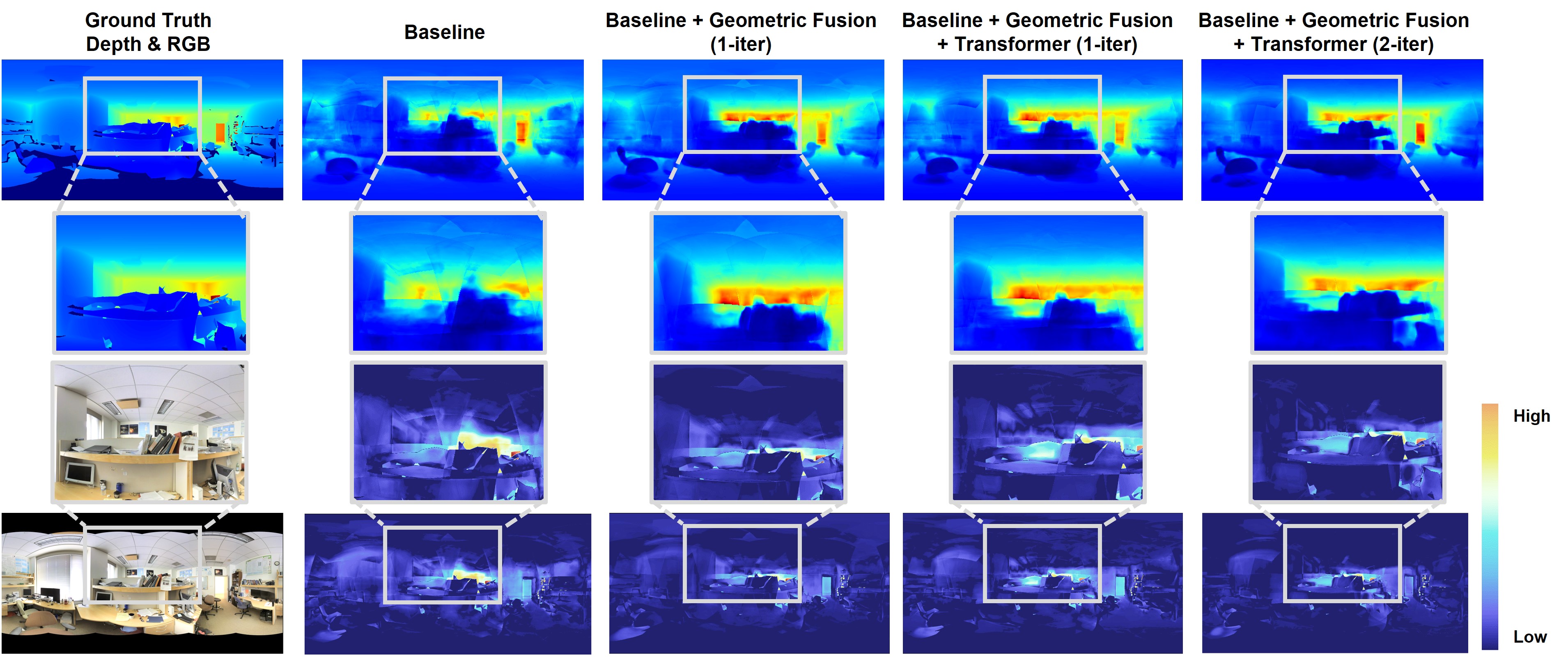}
    \caption{Qualitative comparisons regarding individual components. The top row shows the visual comparisons in depth maps, and the bottom row shows the visual comparisons of the corresponding error maps between the predicted depth maps. The middle two rows show the close-up views of the highlighted areas in the top and bottom rows, respectively. As we add more modules into the pipeline (Figure \ref{fig:pipeline}), the depth estimation becomes more accurate with lower errors, sharper object boundaries and smoother surfaces. The trend of the change in errors can be directly observed from the error maps.
    }
    \label{fig:ab1}
\end{figure*}

\subsection{Overall performance}

We present our model performances and compare it to the existing methods in Table \ref{tab:compare1}. We omit the methods that use supervision signals other than depth \cite{eder2019pano, zeng2020joint} and the self-supervised approaches \cite{zioulis2019spherical} for fair comparison. 
For all datasets, we show our results with 1-iteration (1-iter) and 2-iterations (2-iter). We demonstrate in Table \ref{tab:compare1} that even with 1-iter setting,  our method is able to outperform all the existing methods on Matterport3D \cite{Matterport3D}, and achieve on par performance with current state-of-the-arts on 360D. With 2-iter setting, our method outperforms BiFuse \cite{wang2020bifuse} by $21.4\%$ (Abs Rel) on Stanford2D3D, $56.1\%$ (Abs Rel) on Matterport3D, $30\%$ (Abs Rel) on 360D. Comparing to UniFuse \cite{jiang2021unifuse}, our method improves by  $6.3\%$ (Abs Rel) on Stanford2D3D, $15.3\%$ (Abs Rel) on Matterport3D, $7.7\%$(Abs Rel) on 360D. 
Note that compared to ODE-CNN \cite{cheng2020omnidirectional} which used additional sensor input, our method reduces Abs Rel by $7.9\%$. Qualitative results of our method can be visualized in Figure \ref{fig:quali}. As observed, our method (1-iter and 2-iter) improves the baseline, a direct customization from~\cite{eder2020tangent}, significantly in producing less erroneous depth maps with sharper boundaries and smoother surfaces recovered. 

\subsection{Ablation studies}
\label{sec:ablation}

\firsttable

\noindent \textbf{Individual component study.} We investigate the effectiveness of our method by adding one key component at a time (Table \ref{tab:ab1} and Figure \ref{fig:ab1}). We form our baseline experiment with ResNet34 as encoder without the transformer or the geometric fusion. We experiment on Stanford2D3D, using the configuration of 18 patches, $256\times 256$ patch size, $80^\circ$ FoV. 
As observed from Table \ref{tab:ab1}, the geometry-aware fusion, which only adds less than 2K parameters, is able to improve Abs Rel significantly by $9.7\%$. While being extremely light-weighted, the geometric fusion part proves to be quite beneficial. The incorporation of the transformer, which increases around 19M parameters, leads to another boost of performance by $5.7\%$ (Abs Rel). Together with transformer and geometric fusion, the performance is significantly improved by $15.4\%$ (Abs Rel) with 1-iter setting, and 16.4\% (Abs Rel) with 2-iter setting.  Qualitative results are shown in Figure \ref{fig:ab1}. 
As observed, as we add more modules into our pipeline, the output depth map appears to show fewer artifacts and more structural details. In the meantime, the visualized error maps clearly show the decreasing trend of estimation errors.

\secondtable

\noindent \textbf{Patch size and number of patches.} 
Patch size and the number of patches affect both the accuracy and the efficiency of the method. In this study, we aim to find an optimal balance between efficiency and performance. Theoretically, neither a large patch size nor a large number of patches is desired since they both lead to higher computational complexity. However, table \ref{tab:ab2} also indicates the patch size can not be too small, since the monocular depth estimation requires large-enough FoV to hypothesis the depth scale. We also observe that keep increasing the number of patches (e.g., $>=26$) can degrade the performance, since a larger number of patches also increases the overlapping area, which in turn may intensify the discrepancy problem. 
As a result, we choose to use a relatively small number of patches $N=18$ with a relatively large resolution $256\times256$ to balance between efficiency and performance.

\thirdtable
\noindent \textbf{Image encoder and number of iterations.} We compare the performance of leveraging different image encoders. As listed in Table \ref{tab:ab3}, ResNet34 \cite{he2016deep} outperforms ResNet18 with more complexity. This indicates the potential of our method, as one can incorporate a more sophisticated encoder network. 
We also study the influence of iterations. 
We use the 2-iteration framework for the training since we expect the trained network to handle different types of 3D coordinates. While for testing, we compare 1-4 iterations respectively on the two backbones. 
As seen from Table \ref{tab:ab3}, there is an evident improvement from 1-iter to 2-iter, a slighter improvement from 2-iter to 3-iter, and no gain from 3-iter to 4-iter. Considering the trade-off in performance and the speed, we opt to choose 1-iter or 2-iter settings.



\section{Conclusion}
\label{sec:con}
In this paper, we propose a novel pipeline, \textit{OmniFusion}, for 360 monocular depth estimation. To address the spherical distortion presented in 360 images, as well as to improve the scalability to high-resolution inputs,
we use gnomonic projection-based tangent image presentation. To alleviate the discrepancy between patches, we introduce a geometry-aware fusion mechanism which fuse 3D geometric features with the image features. A self-attention transformer is integrated into our pipeline to globally aggregate information from patches, which leads to more consistent patch-wise predictions. We further extend the geometry-aware fusion with an iterative refining scheme which further improves the depth estimation with more structural details. 
We show that \textit{OmniFusion} effectively mitigates distortion, and significantly improves the depth estimation performance.
Our experiments show that our method achieves state-of-the-art performances on several datasets.

\section*{Acknowledgments}
The research of Yuyan Li and Ye Duan were partially supported by the National Science Foundation under award CNS-2018850, National Institute of Health under awards NIBIB-R03-EB028427 and NIBIB-R01-EB02943, and U.S. Army Research Laboratory W911NF2120275. Any opinions, findings, and conclusions or recommendations expressed in this publication are those of the authors and do not necessarily reflect the views of the U.\,S.\ Government or agency thereof.

\cleardoublepage
{\small
\bibliographystyle{ieee_fullname}
\bibliography{egbib}
}

\clearpage
\thispagestyle{empty}
\appendix
\section{Gnomonic projection}
\label{sec:gnomonic}

\begin{figure}[t]
    \centering
    \includegraphics[width=5cm]{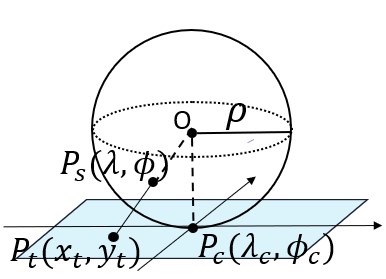}
    \caption{The illustration of the gnomonic projection. A point $P_s(\lambda, \phi)$ located on the spherical sphere is projected onto a point $P_t(x_t, y_t)$ on the flat plane which is tangent to a point $P_c(\lambda_c, \phi_c)$. 
     }
    \label{fig:projection}
\end{figure}

We use the distortion-free tangent image representation to address the irregular 360 image distortion. Tangent image is the \textit{gnomonic projection} of a sphere surface onto a flat, rectangular plane surface.
The gnomonic projection \cite{coxeter1961introduction} (Figure \ref{fig:projection}) is a map projection obtained by projecting points $P_s$ on the surface of sphere from a sphere's center $O$ to point $P_t$ in a plane that is tangent to a point $P_c$. 

For a pixel on the ERP image $P_e(x_e, y_e)$, we first find its corresponding point $P_s(\lambda, \phi)$ locating on the unit sphere. 
\begin{equation}
    \lambda = \frac{2\pi x_e}{W},\; \phi = \frac{\pi y_e}{H}
\end{equation}
where $H$ and $W$ are height and width of the ERP image.
The projection from $P_s(\lambda, \phi)$ to $P_t(x_t, y_t)$ is defined as:
\begin{equation}
\begin{aligned}
    &x_t = \frac{\mathbf{cos}(\phi)\mathbf{sin}(\lambda - \lambda_c)}{\mathbf{cos}(c)} \\
    &y_t = \frac{\mathbf{cos}(\phi_c)\mathbf{sin}(\phi) - \mathbf{sin}(\phi_c)\mathbf{cos}(\phi)\mathbf{cos}(\lambda - \lambda_c)}{\mathbf{cos}(c)} \\
    &\mathbf{cos}(c) = \mathbf{sin}(\phi_c)\mathbf{sin}(\phi) + \mathbf{cos}(\phi_c) \mathbf{cos}(\phi) \mathbf{cos}(\lambda - \lambda_c)
\end{aligned}
\label{eq:sphere2xy}
\end{equation}
where $(\lambda_c, \phi_c)$ are the spherical coordinates of the tangent plane center $P_s$.

The inverse gnomonic transformations are:
\begin{equation}
\begin{aligned}
    &\lambda = \lambda_c + \mathbf{tan}^{-1} (\frac{x_t\;\mathbf{sin}(c)}{\gamma\;\mathbf{cos}(\phi_1)\mathbf{cos}(c) - y_t\;\mathbf{sin}(\phi_c)\mathbf{sin}(c)}) \\
    &\phi = \mathbf{sin}^{-1} (\mathbf{cos}(c)\mathbf{sin}(\phi_c) + \frac{1}{\gamma} y_t \mathbf{sin}(c)\mathbf{cos}(\phi_c)) 
\end{aligned}
\label{eq:xy2sphere}
\end{equation}
where $\gamma = \sqrt{x_t^2 + y_t^2}$ and $c=\mathbf{tan}^{-1}\gamma$.

With Equation \ref{eq:sphere2xy} and \ref{eq:xy2sphere}, we can build one-to-one forward and inverse mapping functions between pixels on the ERP image and pixels on the tangent image.

\section{Geometry-aware feature fusion}

\begin{figure*}
    \centering
    \includegraphics[width=16cm]{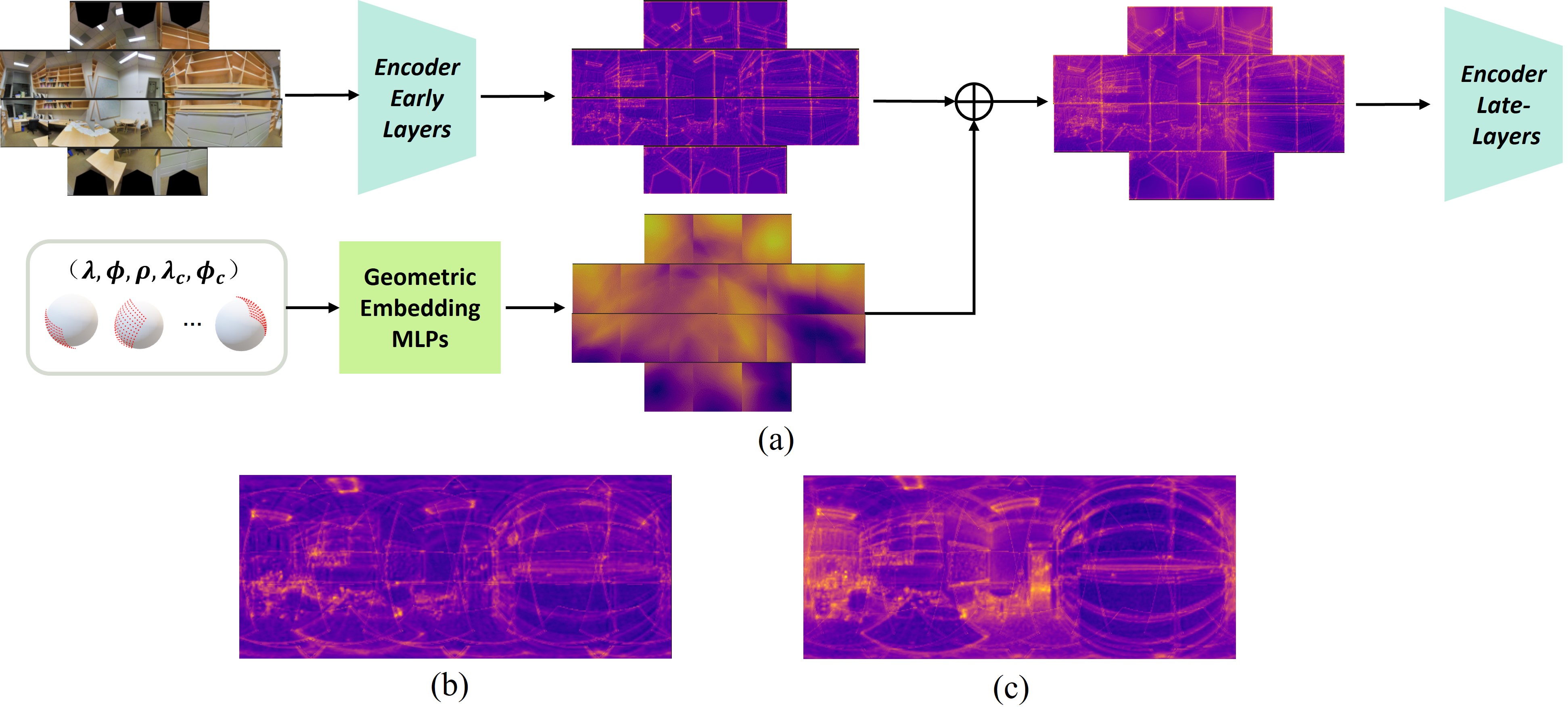}
    \caption{(a) Detailed pipeline of geometry-aware feature fusion. A set of tangent images are encoded into a set of image feature maps, while the 3D coordinates are encoded and converted into a set of geometric feature maps. The patch-wise 2D image features are fused with the patch-wise geometric feature.  (b) The merged ERP feature map of patch features without the geometric fusion. (c) The merged ERP feature map of patch features with the geometric fusion. Comparing to the merged ERP feature maps without geometric fusion in (b), the geometry-aware fused ERP feature map in (c) appears to be more locally consistent.}
    \label{fig:geo:fuse:illus1}
\end{figure*}

As the geometry-aware feature fusion module is one of the major innovations of our paper, in this section we provide more detailed illustrations. As shown in Figure~\ref{fig:geo:fuse:illus1}, more intermediate representations involved in the module is visualized. Specifically, the patch-wise 2D image features and the patch-wise geometric features are visualized separately, along with the feature maps after fusion, in which the mean value of each feature is shown. For visual comparison, the patch-wise features before Figure~\ref{fig:geo:fuse:illus1} (b) and after fusion (c) are projected and merged into two ERP feature maps. As observed, the fused feature maps inherit more locally consistent structures, which is expected to lead to more locally consistent depth results. It is worth  mentioning that patch-wise geometric features are fixed once learned when the inputs are just based on the spherical coordinates with fixed $\rho$, and independent from the image. This means no extra computation in inference is needed for the first iteration. While for the second iteration, since $\rho$ depends on the input image, new geometric features need to be re-computed, but the MLPs are super light-weight compared to the original CNNs. 

\begin{figure*}
    \centering
    \includegraphics[width=14cm]{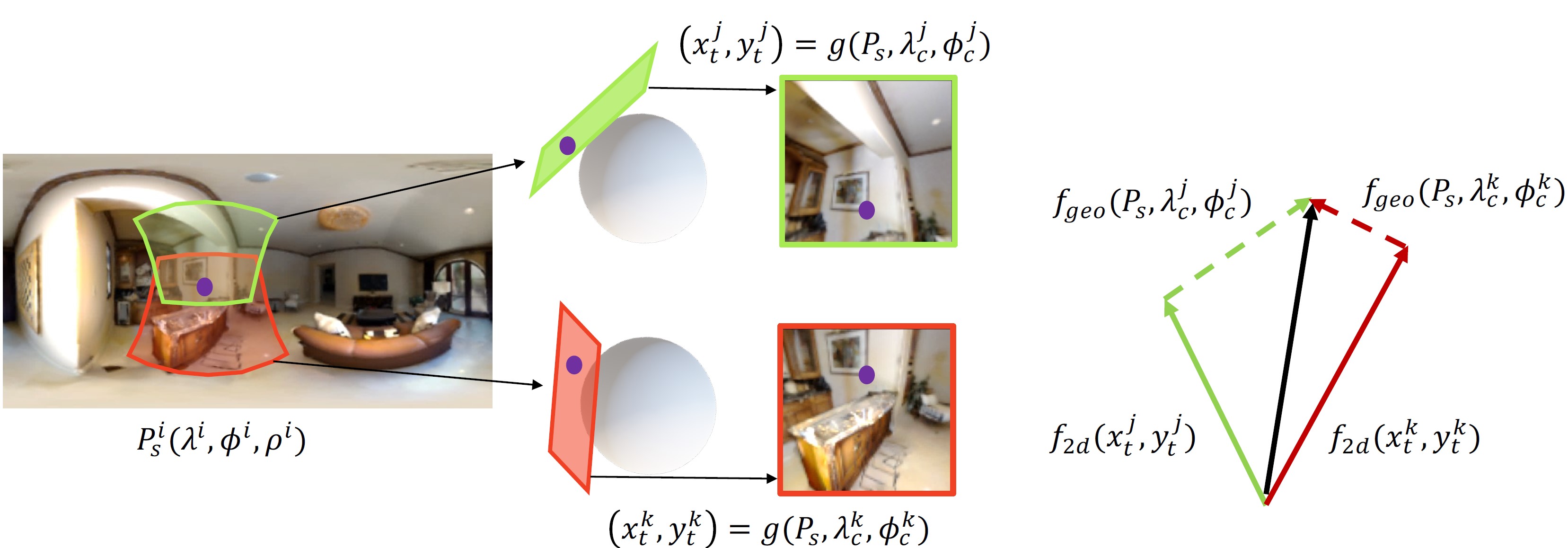}
    \caption{A more intuitive view of geometry-aware feature fusion. Based on the gnomonic geometry, a single point from the ERP space, $P_s^i (\lambda^i, \phi^i, \rho^i)$ is projected to two tangent images centered at $(\lambda_c^j, \phi_c^j)$ and $(\lambda_c^k, \phi_c^k)$, and appear at two different pixels $(x_t^j, y_t^j)$ and $(x_t^k, y_t^k)$, respectively. Image features located at the two pixels can be visualized in high-dimensional vectors (solid green and red arrows in the right panel, respectively). Since the discrepancy is caused by the gnomonic transformation from $(P_s, \lambda_c, \phi_c)$, we utilize geometric features encoded from $(P_s, \lambda_c, \phi_c)$ to compensate for the discrepancy (dashed arrows).}
    \label{fig:geo:fuse:illus2}
\end{figure*}

The intuition behind the geometry-aware fusion design can be visualized in high-dimensional feature space, see Figure~\ref{fig:geo:fuse:illus2}. Based on the Equation~\ref{eq:sphere2xy}, a single point from the ERP space, $P_s^i (\lambda^i, \phi^i, \rho^i)$, is projected to two tangent images centered at $(\lambda_c^j, \phi_c^j)$ and $(\lambda_c^k, \phi_c^k)$, and appear at $(x_t^j, y_t^j)$ and $(x_t^k, y_t^k)$, respectively. As observed, different appearances at the two points can lead to different image features encoded from the shared CNN kernel, which can be visualized as high-dimensional vectors (solid green and red arrows on the right panel). Such difference in the 2D features will make the merged results appear to be locally inconsistent. Since the discrepancy is caused by the gnomonic transformation from $(P_s, \lambda_c, \phi_c)$, we believe a point-encoding model can learn a geometric embedding space out of $(P_s, \lambda_c, \phi_c)$ to mitigate the discrepancy (dashed arrows). While $P_s$ makes the embedding to be aware of the global position, $(\lambda_c, \phi_c)$ differentiates between patches to enable the compensation. 

\begin{table*}[hbt!]
\begin{center}
\begin{footnotesize}
\begin{tabular}{b{12em} b{4em} | b{4em}  b{4em} b{4em}b{5em} b{4em} b{4em} b{4em}  } 
\hline 

\textbf{Configurations}  & \textbf{\#Params} & \textbf{Abs Rel}$\downarrow$ &\textbf{ Sq Rel} $\downarrow$ & \textbf{RMSE}$\downarrow$ & \textbf{RMSE(log)}$\downarrow$ & $\pmb{\delta_1}$  $\uparrow$ & $\pmb{\delta_2}$  $\uparrow$ & $\pmb{\delta_3}$  $\uparrow$ \\
\hline
depth = 2, num of heads = 2	&	19M	 & 0.1091 & 0.0614	 & 0.3885  & 0.1782  & 0.8738	 & 0.9670 &	0.9891 \\
depth = 4, num of heads = 4	&   24M	 &	\textbf{0.1016} & 0.0583 &	\textbf{0.3796} & 0.1774 & 0.8867 &	0.9688 &	0.9885\\
depth = 6, num of heads = 4 & 	32M	 & 0.1026 &	\textbf{0.0572} & 0.3883 &\textbf{0.1753}	& \textbf{0.8893}	& \textbf{0.9689} &	\textbf{0.9892}\\
depth = 8, num of heads = 8 &	38M	 & 0.1044 & 0.0596 & 0.3926 & 0.1819 &	0.8739 &	0.9650 &	0.9873
\\
\hline
\end{tabular}
\end{footnotesize}
\end{center}
\caption{The ablation study of the transformer configurations. We use ResNet18 as encoder for all experiments.}
\label{tab:transformer}
\end{table*}

\section{Transformer Architecture and Ablation Study}

The architecture of the multi-head attention transformer  follows \cite{vaswani2017attention}:
\begin{equation}
    \begin{aligned}
        z_0 &= [x^1 E, x^2 E, ..., x^N E] + E_{pos}, \\
        z_l' &= Norm(MSA(z_{l-1}, z_0) + z_{l-1}), \\
        z_l &= Norm(FFN(z_l') + z_l'),
    \end{aligned}
\label{eq:transformer}
\end{equation}
where $Norm$ represents layer normalization, $l = 1, ..., L$ is the index of the transformer block. 
The multi-headed self-attention (MSA)  is computed as:
\begin{equation}
\begin{aligned}
    &MSA(X) =  concat_{h=1}^H [Attn_h(X)]W  \\
    &Attn_h(X) = softmax(\frac{QK^T}{\sqrt{d_h}})V \\
    &Q=XW_Q, \; K=XW_K, \;, V=XW_V
\end{aligned}
\end{equation}
where $Q, K, V$ correspond to query, key, value matrix, respectively. $h$ denotes the number of heads. We reshape the transformer output, then use another $1\times 1$ convolution layer to increase feature dimension, and add the encoder output as residual. 

An ablation study on the transformer depth and the number of heads is shown in Table \ref{tab:transformer}. The ablation study here is conducted based on ResNet18, not the ResNet34  used in our final pipeline, in order to conduct the experiments more efficiently. The number of parameters shown in the table considers the entire network rather than the transformer module alone. We chose 6 transformer blocks (depth=6) and a number of 4 heads (number of heads=4) as the default configuration, as this configuration tends to have fewer errors and higher inlier ratios. 

\section{Loss Function}
Our network is trained in an end-to-end fashion. We adopt BerHu loss \cite{laina2016deeper} for optimizing depth predictions of all iterations. 
\begin{equation}
    \mathcal{L}_{depth}  = \left\{\begin{matrix}
    |\Delta D|, |\Delta D| \leq c \\ 
    \frac{\Delta D^2 + c^2}{2c}, |\Delta D| > c
\end{matrix}\right.
\end{equation}
where $\Delta D = |D_{gt} - D_e| * M$ is the absolute difference of ground truth depth $D_{gt}$ and the predicted depth $D_e$. $M$ is a binary mask that mask out invalid depth pixels. $c$ is a border value defined as the 20\% of the maximum per batch residual $c = 0.2max(\Delta D)$.

The final loss term is the combination of losses from all iterations:
\begin{equation}
    \mathcal{L}_{total} = \sum_i \mathcal{L}_{depth}
\end{equation}

\section{Generalization}
We conducted a cross-dataset evaluation and summarized the results in Table \ref{tab:general}. All methods in the table are trained on Matterport3D \cite{Matterport3D} training set and evaluated on Stanford2D3D \cite{armeni2017joint} test set. We used the official pre-trained models and the evaluation code provided by UniFuse \cite{jiang2021unifuse} and HoHoNet \cite{sun2021hohonet} for a fair comparison. As observed, our method showed superior generalization ability compared to these state-of-the-arts methods. 
\vspace{-0.1cm}
\begin{table}[h]
    \centering
    \begin{footnotesize}
    \begin{tabular}{c|c|c|c}
    \hline
        Methods & Abs Rel$\downarrow$ & Sq Rel$\downarrow$ & RMSE$\downarrow$  \\
    \hline
        UniFuse \cite{jiang2021unifuse} & 0.1192 & 0.0813 & 0.4291 \\
        HoHoNet \cite{sun2021hohonet} & 0.1083 & 0.0755 & 0.4166\\
    \hline
        \textbf{OmniFusion, Ours}  & \textbf{0.1044} & \textbf{0.0620} & \textbf{0.3781}\\
    \hline
    \end{tabular}
    \end{footnotesize}
    \caption{Cross-dataset evaluation.}
    \label{tab:general}
\end{table}

\section{Additional qualitative  comparisons}
\label{sec:visual}

\begin{figure*}
    \centering
    \includegraphics[width=17cm]{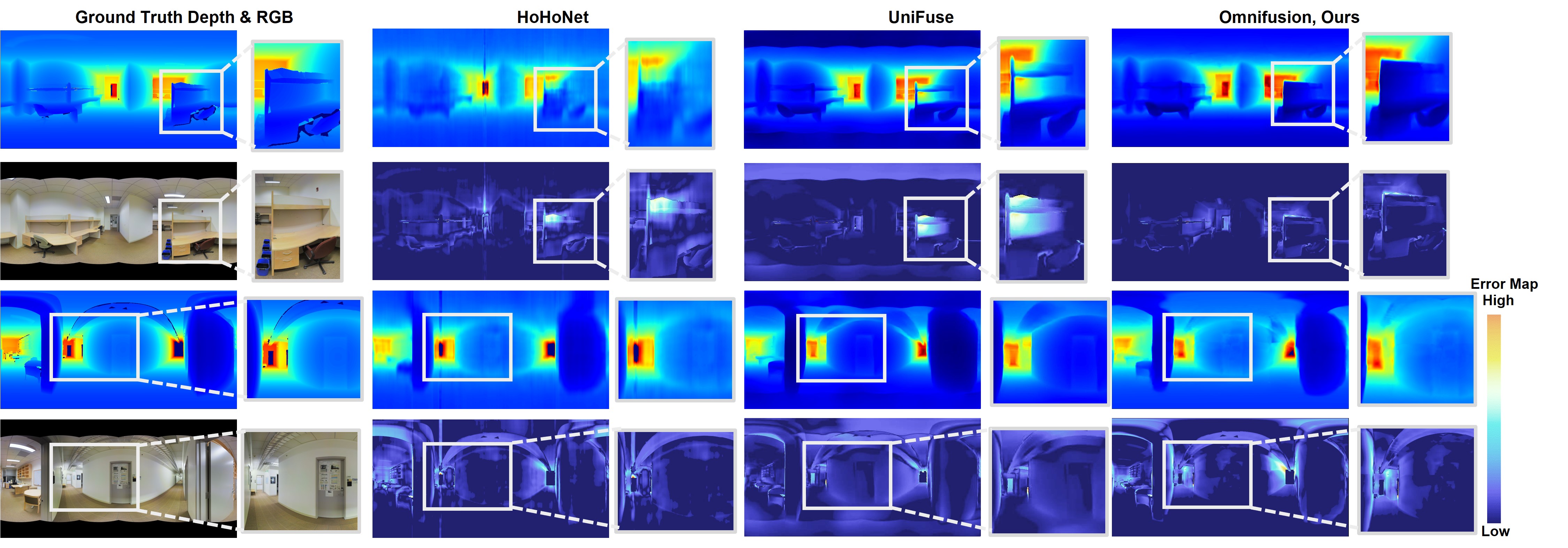}
    \caption{The qualitative comparisons with the current state-of-the-art works on the dataset Stanford2D3D \cite{armeni2017joint}. We show the results of HoHoNet \cite{sun2021hohonet} (second column), UniFuse \cite{jiang2021unifuse} (third column), and ours (last column). Both the depth maps and the error maps against the ground-truth are included for comparison. See the zoomed-in areas for detailed comparisons. }
    \label{fig:comp_st}
\end{figure*}

\begin{figure*}
    \centering
    \includegraphics[width=17cm]{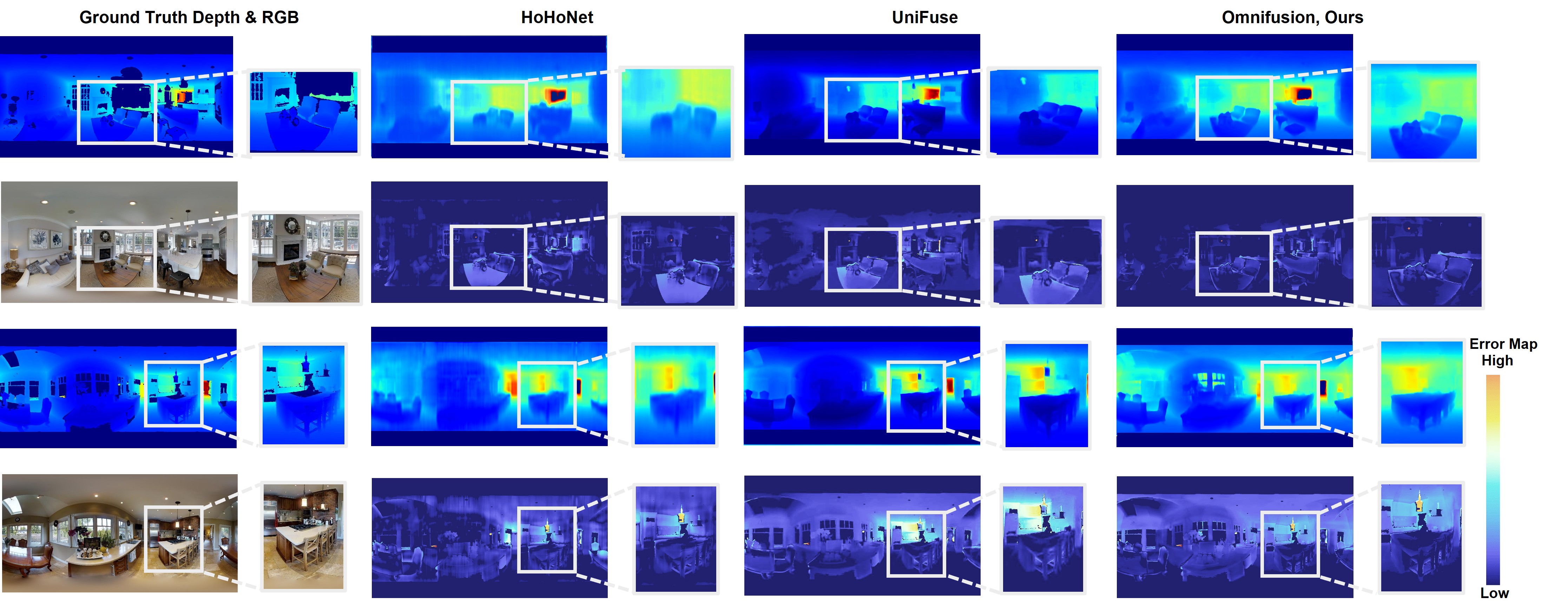}
    \caption{The qualitative comparisons with current state-of-the-art works on the dataset Matterport3D \cite{Matterport3D}. We show the results of HoHoNet \cite{sun2021hohonet} (second column), UniFuse \cite{jiang2021unifuse} (third column), and ours (last column).  Both the depth maps and the error maps against the ground-truth are included for comparison. See the zoomed-in areas for detailed comparisons.}
    \label{fig:comp_mp3d}
\end{figure*}

\begin{figure*}
    \centering
    \includegraphics[width=16cm]{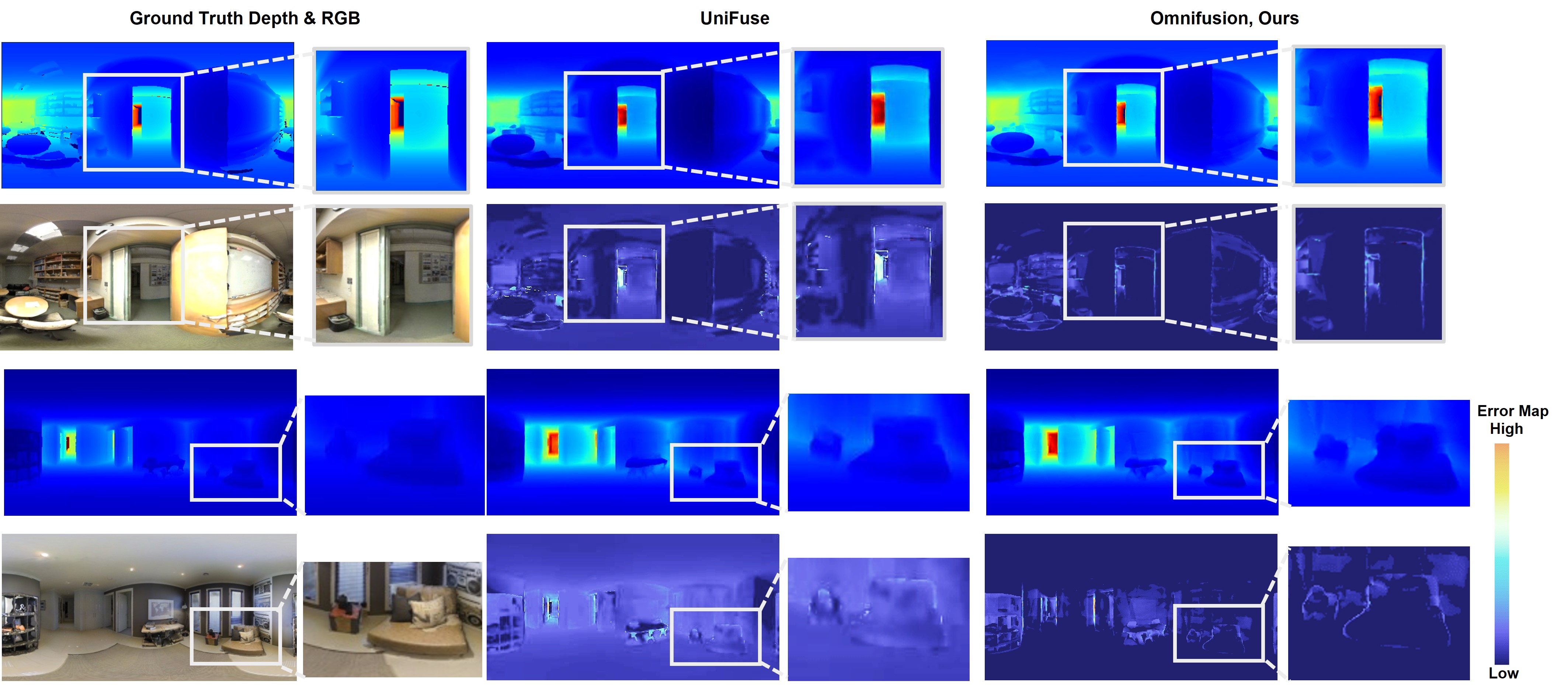}
    \caption{The qualitative comparisons with current state-of-the-art works on the dataset 360D \cite{zioulis2018omnidepth}, We show the results of UniFuse  \cite{jiang2021unifuse} (second column), and ours (last column). Both the depth maps and the error maps against the ground-truth are included for comparison. See the zoomed-in areas for detailed comparisons.}
    \label{fig:comp_360d}
\end{figure*}

\begin{figure*}
    \centering
    \includegraphics[width=16cm]{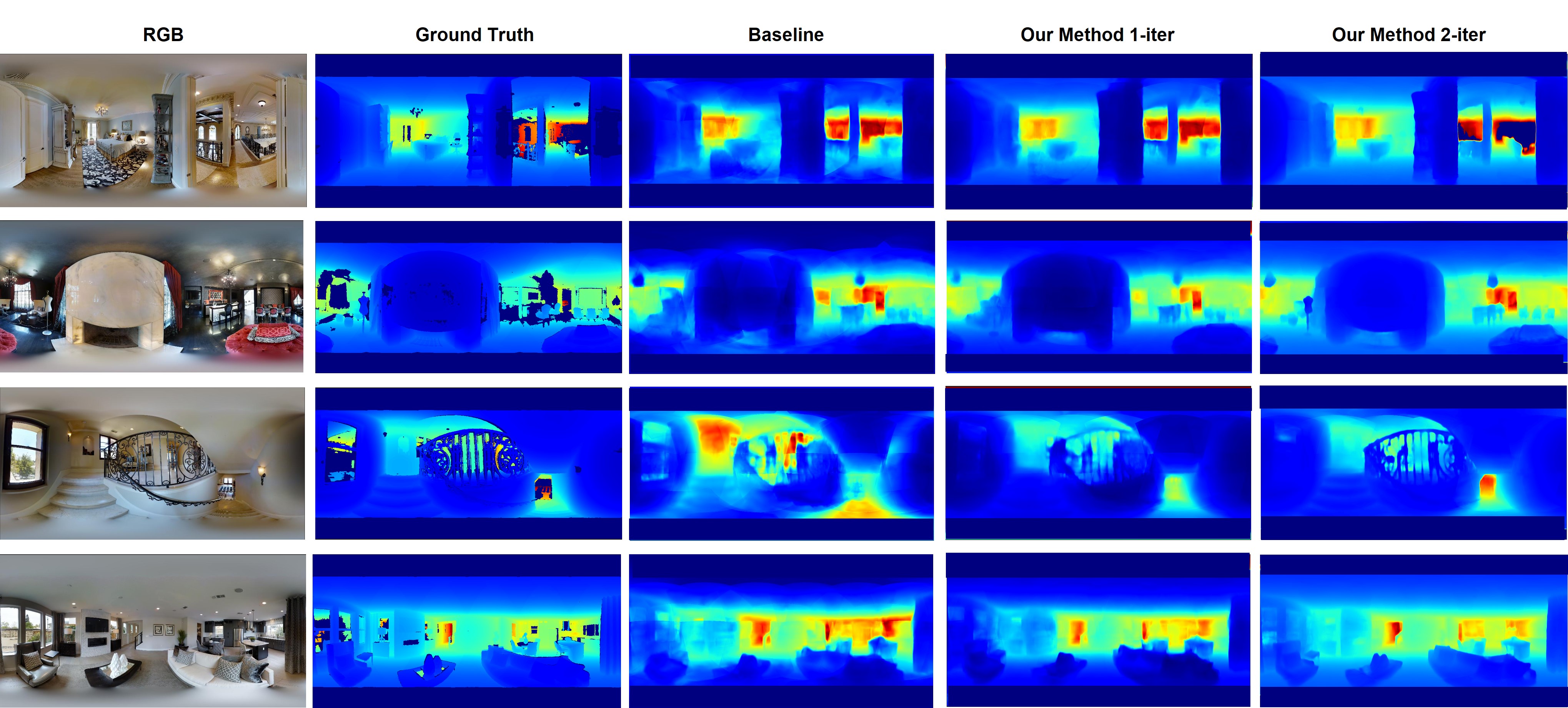}
    \caption{More qualitative results of OmniFusion on Matterport3D \cite{Matterport3D}. }
    \label{fig:mp3d_ext}
\end{figure*}

Besides the qualitative comparison between our method and the baseline method tailored from~\cite{eder2020tangent}, we also extend to qualitatively compare our method with current state-of-the-art methods, HoHoNet \cite{sun2021hohonet} and UniFuse \cite{jiang2021unifuse} on three datasets: Stanford2D3D \cite{armeni2017joint}, Matterport3D \cite{Matterport3D}, and 360D \cite{zioulis2018omnidepth}. The results are shown in Figure \ref{fig:comp_st}, \ref{fig:comp_mp3d}, \ref{fig:comp_360d}, respectively. We use the pretrained models downloaded from their official GitHub repositories, respectively. \footnote{https://github.com/sunset1995/HoHoNet}
\footnote{https://github.com/alibaba/UniFuse-Unidirectional-Fusion} Note that the results from HoHoNet~\cite{sun2021hohonet} are not included in Figure~\ref{fig:comp_360d} because they have not reported results or releases code on 360D \cite{zioulis2018omnidepth} dataset. Figure \ref{fig:mp3d_ext} shows additional qualitative results of our OmniFusion on Matterport3D \cite{Matterport3D} besides what have been  provided on Stanford2D3D \cite{armeni2017joint} in the main paper.
All of these comparisons clearly show that our method recovers more structural details in the final depth maps, maintains sharp edges, smooth surfaces, and exhibits fewer errors.


\end{document}